# Gaussian Mixture Reduction for Time-Constrained Approximate Inference in Hybrid Bayesian Networks

Cheol Young Park, Kathryn Blackmond Laskey, Paulo C. G. Costa, Shou Matsumoto
The Sensor Fusion Lab & Center of Excellence in C4I
George Mason University, MS 4B5
Fairfax, VA 22030-4444 U.S.A.
cparkf@masonlive.gmu.edu, [klaskey, pcosta]@gmu.edu, smatsum2@masonlive.gmu.edu

*Abstract*— **Hybrid Bayesian Networks (HBNs), which contain both discrete and continuous variables, arise naturally in many application areas (e.g., image understanding, data fusion, medical diagnosis, fraud detection). This paper concerns inference in an important subclass of HBNs, the conditional Gaussian (CG) networks, in which all continuous random variables have Gaussian distributions and all children of continuous random variables must be continuous. Inference in CG networks can be NP-hard even for special-case structures, such as poly-trees, where inference in discrete Bayesian networks can be performed in polynomial time. Therefore, approximate inference is required. In approximate inference, it is often necessary to trade off accuracy against solution time. This paper presents an extension to the Hybrid Message Passing inference algorithm for general CG networks and an algorithm for optimizing its accuracy given a bound on computation time. The extended algorithm uses Gaussian mixture reduction to prevent an exponential increase in the number of Gaussian mixture components. The trade-off algorithm performs pre-processing to find optimal run-time settings for the extended algorithm. Experimental results for four CG networks compare performance of the extended algorithm with existing algorithms and show the optimal settings for these CG networks.**

## I. INTRODUCTION

A Bayesian Network (BN) [Pearl, 1988] is a probabilistic graphical model that represents a joint distribution on a set of random variables in a compact form that exploits conditional independence relationships among the random variables. The random variables (RVs) are represented as nodes in a directed acyclic graph (DAG) in which a directed edge represents a direct dependency between two nodes and no directed cycles are allowed in the graph. Bayesian Networks have become a powerful tool for representing uncertain knowledge and performing inference under uncertainty. They have been applied in many domains, such as Image Understanding, Data fusion, Medical diagnosis, and Fraud detection, and have become a powerful tool in inference for the real world.

Hybrid Bayesian Network (HBNs) can contain both discrete and continuous RVs. An important subclass, the conditional Gaussian (CLG) networks, consists of networks in which all discrete random variables have only discrete parents, all continuous random variables have Gaussian distributions, and the conditional distribution of any Gaussian RV is linear in its Gaussian parents. Exact inference methods exist for CLG networks [Lauritzen, 1992][Lauritzen & Jensen, 2001]. However, even in special cases for which exact inference in discrete Bayesian Networks (BNs) is tractable, exact inference in CLG networks can be NP-hard [Lerner & Parr, 2001]. In particular, the posterior marginal distribution for each individual Gaussian random variable is a mixture of Gaussian distributions, and the number of components needed to compute the exact distribution for a given random variable may be exponential in the number of discrete variables in the network. Furthermore, no exact algorithms exist for general CG networks. Therefore, approximate inference for CG networks is an important area of research.

Approximate inference algorithms for HBNs can be roughly classified into five categories: (1) Sampling (SP), (2) Discretization (DS), (3) Structure Approximation (SA), (4) Clustering (CL), and (5) Message Passing (MP) approaches.

SP algorithms draw random samples to use for inference and can handle BNs of arbitrary structure. Henrion [1988] presented a basic sampling approach, called logic sampling, for approximate inference in discrete Bayesian networks. Logic sampling generates samples beginning at root nodes and following links to descendant nodes, terminating at leaf nodes of the graph. If a sampled realization contains an evidence node whose value does not match the observed value, it is rejected. The result is a sample from the conditional distribution of the sampled nodes given the observed values of the evidence nodes. This rejection strategy may require a very large number of samples to converge to an acceptable inference result. Further, this strategy cannot be applied when there are continuous evidence nodes, because in this case all samples would be rejected. Fung & Chang [1989] suggested a method that sets all evidence variables to their observed values, samples only the non-evidence variables, and weights each sample by the likelihood of the evidence variables given the sampled non-evidence variables. This likelihood weighting algorithm, which can be applied when evidence nodes are continuous, has become very popular. However, when the evidence configuration is highly unlikely, this method can result in very poor accuracy. Pearl [1988] proposed a Gibbs sampling approach for Bayesian networks. His algorithm is a specific case of the more general class of Markov Chain Monte Carlo algorithms [Gilks et al, 1996]. Efficiency of Gibbs sampling can be dramatically improved by sampling only a subset (called a *cutset*) of random variables that breaks all loops in the graph, and performing exact inference on the remaining singly connected network [Bidyuk & Dechter, 2007]. Nevertheless, for any SP algorithm very large numbers of samples may be required for challenging BNs, such as those with complex topologies, very unlikely evidence configurations, and/or deterministic or near-deterministic relationships.

DS algorithms change a hybrid BN to a discrete BN by discretizing all continuous RVs in the hybrid BN. This approach changes a continuous variable to a set of intervals, called a bin. After the change, the discretized BN is handled by a discrete inference algorithm (e.g., [Pearl, 1988]). Kozlov & Koller [1997] provided an improved discretization by efficiently adjusting the shape of a continuous RV. However, DS algorithms start with approximation for discretization and this approximation can cause inaccurate posterior distributions. Accuracy can be improved with finer discretization, but at the cost of possibly major additional cost in time and space. Furthermore, there is a time cost for discretization, and a need for methods to choose the granularity of the distribution to balance accuracy against computation cost.

SA algorithms change an intractable hybrid BN (e.g., conditional nonlinear Gaussian network) to a tractable hybrid BN (e.g., conditional linear Gaussian network). After changing to a tractable hybrid BN, a hybrid inference algorithm which can handle the tractable hybrid BN is used for inference. Shenoy [2006] proposed a SA algorithm in which any type of a continuous RV can be approximated by a mixture of Gaussian distributions, thus converting an arbitrary hybrid BN to a CG BN. He showed how various hybrid BNs (e.g., non-Gaussian HBN, nonlinear HBN, a HBN with a continuous parent and a discrete child, and a HBN with non-constant variance) can be converted to a CG BN. Although SA algorithms can treat various types of HBNs, they require an appropriate CG inference algorithm for the converted HBN.

CL algorithms handle loops by converting the original BN to a graph of clusters in which each node corresponds to a cluster of nodes from the original BN, such that the graph of clusters contains no loops. A conversion step is required to form clusters from the original BN. Among CL approaches, the popular Junction Tree (JT) algorithm has been adapted for inference in CG networks [Lauritzen, 1992][Lauritzen & Jensen, 2001]. However, constraints required by the Lauritzen algorithm [Lauritzen, 1992][Lauritzen & Jensen, 2001] on the form of the junction tree tend to result in cliques containing very many discrete nodes. Because inference is exponential in the number of discrete nodes in a cluster, the algorithm is often intractable even when a tractable clustering approach exists for a discrete network of the same structure [Lerner & Parr, 2001]. For this reason, it is typically necessary to resort to approximate inference. Gaussian mixture reduction (GMR) has been suggested as an approximation approach [Lerner, 2002]. GMR approximates a Gaussian mixture model (GMM) with a GMM having fewer components.

In MP algorithms, each node in the BN sends messages to relevant nodes along paths between the relevant nodes. The messages contain information to update the distributions of the relevant nodes. After updating, each of the nodes computes its marginal distribution. If the BN has loops, message passing may not converge. MP algorithms are also subject to the problem of uncontrolled growth in the number of mixture components. A GMR approach has been proposed to address this issue [Sun et al., 2010][Sun & Chang, 2010]. However, they provided no general algorithm for applying GMR within the MP algorithm. Park et al. [2015] introduced a general algorithm for MP using GMR[1], but included no guidance on how to trade off between accuracy and computational resources in hybrid MP using GMR.

This paper presents a complete solution to the hybrid inference problem by providing two algorithms: Hybrid Message Passing (HMP) with Gaussian Mixture Reduction (GMR) and Optimal Gaussian Mixture Reduction (Optimal GMR).

The HMP-GMR algorithm prevents exponential growth of Gaussian mixture components in MP algorithms for inference in CG Bayesian networks. We present an extension of the algorithm of [Sun et al., 2010][Sun & Chang, 2010] that incorporates GMR to control complexity, and examine its performance relative to competing algorithms.

Each inference algorithm has its own characteristics. For example, some algorithms are faster and some are more accurate. Further, accuracy and speed can depend on the Bayesian network and the specific pattern of evidence. These characteristics can be used as guidance for choosing an inference method for a given problem. Metrics for evaluating an inference algorithm include speed, accuracy, and resource usage (e.g., memory or CPU usage). In some situations, algorithm speed is the most important factor. In other cases, accuracy may be more important. For example, early stage missile tracking may require a high speed algorithm for estimating the missile trajectory, while matching faces in a security video against a no-fly database may prioritize accuracy over speed. The HMP-GMR algorithm requires a maximum number of Gaussian components as an input parameter. This maximum number of components influences both accuracy and execution time of the HMP-GMR algorithm. We introduce a preprocessing algorithm called HMP-GMR with Optimal Settings (HMP-GMR-OS), which optimizes the initial settings for HMP-GMR to provide the best accuracy on a given HBN under a bound on computation time. The HMP-GMR-OS algorithm is intended for cases in which a given HBN will be used repeatedly in a time-limited situation, and a pre-processing step is desired to balance accuracy against speed of inference. Sampling approaches have been used for such situations, because of their anytime property. That is, sampling always provides an answer even if it runs out of time. In some cases, our algorithm can result in better accuracy than a sampling approach for the same execution time.

---

[1] This paper is an extension of the conference paper, [Park et al., 2015].

The layout of this paper is as follows. Section 2 introduces Hybrid Message Passing Inference and Gaussian Mixture Reduction. Section 3 presents the HMP-GMR algorithm, which combines the two methods introduced in Section 2. Section 4 proposes the OCB algorithm to find the optimal number of allowable components in any given mixture distribution. Section 5 presents experimental results on the advantages and disadvantages of the new algorithm. Section 6 draws conclusions.

## II. PRELIMINARIES

In this section, we introduce message passing inference for CG BNs and component reduction for Gaussian mixture models.

### A. Hybrid Message Passing Inference

#### Structure of Hybrid Bayesian Network

A general hybrid BN can contain both discrete and continuous nodes. A node in a hybrid BN can be categorized according to its type (i.e., discrete or continuous), its parent node type(s) (i.e., discrete, continuous, or hybrid with at least one discrete and one continuous node), and its child node type(s) (i.e., discrete, continuous, or hybrid). The following table shows all possible classifications of nodes in a hybrid BN (D stands for discrete; C stands for continuous; and H stands for hybrid).

| | Parent node type(s) | D | | | C | | | H | | |
|---|---|---|---|---|---|---|---|---|---|---|
| | Child node type(s) | D | C | H | D | C | H | D | C | H |
| Node type | D | 1 | 2 | 3 | 4 | 5 | 6 | 7 | 8 | 9 |
| | C | 10 | 11 | 12 | 13 | 14 | 15 | 16 | 17 | 18 |

Table 1. Possible Node Types in a Hybrid BN

As shown in Table 1, there are 18 node categories in a general hybrid BN. Various special cases impose restrictions eliminating some of the 18 categories. A hybrid BN in which no discrete node may have a continuous parent node is called a conditional hybrid BN [Sun, 2007]. That is, a conditional hybrid BN may contain Types 1, 2, 3, 11, 14, and 17 from Table 1. These six cases are shown in Fig. 1 below. In the figure, a rectangle indicates a discrete node and a circle indicates a continuous node. For example, Type 1 has a discrete node $B$ with its discrete parent node $A$ and discrete child node $C$, while Type 3 has a discrete node $B$ with its discrete parent node $A$ and hybrid child nodes $C$ and $Y$.

A general hybrid BN places no restriction on the type of probability distribution for a continuous node. If all continuous nodes in a hybrid BN have Gaussian probability distributions, the BN is called Gaussian hybrid BN. A BN that is both a conditional hybrid BN and a Gaussian hybrid BN is called a conditional Gaussian (CG) BN. CG BNs can be further classified into two sub-categories: conditional linear Gaussian (CLG) BNs and conditional nonlinear Gaussian (CNG) BNs. For the CLG BNs, the Gaussian conditional distributions are always linear functions of the Gaussian parents. That is, if $X$ is a continuous node $X$ with $n$ continuous parents $U_1, \ldots, U_n$ and $m$ discrete parents $A_1, \ldots, A_m$, then the conditional distribution $p(X \mid u, a)$ given parent states $U=u$ and $A=a$ has the following form:

$$p(X \mid u, a) = \mathcal{N}(L^{(a)}(u), \sigma^{(a)}), \qquad (1)$$

where $L^{(a)}(u) = m^{(a)} + b_1^{(a)} u_1 + \cdots + b_n^{(a)} u_n$ is a linear function of the continuous parents, with intercept $m^{(a)}$, coefficients $b_i^{(a)}$, and standard deviation $\sigma^{(a)}$ that depend on the state $a$ of the discrete parents.

A Gaussian conditional distribution for a continuous node in a CNG BN can be any function of the Gaussian and discrete parents. The form is similar to Equation 1 except that $L^{(a)}(u)$ can be a nonlinear function.

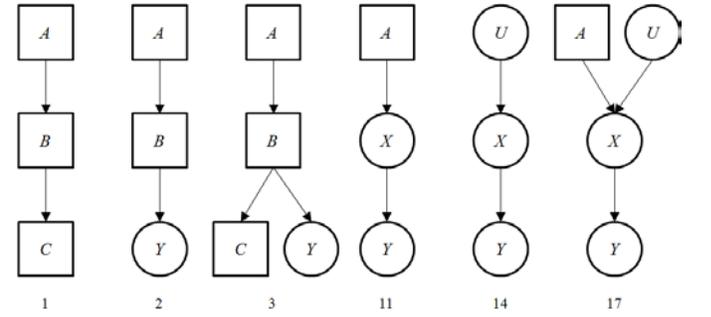

Fig. 1. Possible Node Types for a Conditional Gaussian Hybrid BN (rectangle denotes discrete node; circle denotes continuous node)

Note that the types of Fig. 1 cover any number of parent and child nodes. Thus, the discrete node $B$ can have a set of discrete parent nodes (i.e., $\mathbf{A} = \{A_1, A_2, ..., A_l\}$), a set of discrete child nodes (i.e., $\mathbf{C} = \{C_1, C_2, ..., C_m\}$), and/or a set of continuous child nodes (i.e., $\mathbf{Y} = \{Y_1, Y_2, ..., Y_n\}$). The continuous node $X$ can have a set of discrete parent nodes (i.e., $\mathbf{A} = \{A_1, A_2, ..., A_l\}$), a set of continuous parent nodes (i.e., $\mathbf{U} = \{U_1, U_2, ..., U_m\}$), and a set of continuous child nodes (i.e., $\mathbf{Y} = \{Y_1, Y_2, ..., Y_n\}$). We use this notation to introduce message passing inference for a discrete BN, a continuous BN, and a hybrid BN.

#### Message Passing Inference for Discrete BN

Message passing inference for a discrete BN was introduced in [Pearl, 1988]. A discrete BN contains only discrete nodes (i.e., Type 1 in Fig. 1). The objective of inference is to compute the function

$$\text{BEL}(B) = P(B \mid e) \qquad (2)$$

for each node $B$ in the Bayesian network. Here, $B$ denotes a node with its associated RV, $e$ is a set of evidence events consisting of state assignments for nodes in the network, and $\text{BEL}(B)$ is the conditional distribution of $B$ given the values of the evidence RVs. If the BN is a polytree (has no undirected cycles), the evidence $e$ can be split into two components, $e_B^+$ and $e_B^-$, where $e_B^+$ relates only to non-descendants of $B$ and $e_B^-$ relates only to descendants of $B$. Equation 2 can be decomposed as follows:

$$\begin{aligned} P(B \mid e) &= \alpha P(B \mid e_B^+, e_B^-) \\ &= \alpha P(B \mid e_B^+) P(B \mid e_B^-) \\ &= \alpha \pi(B) \lambda(B) \end{aligned}$$

where $\alpha$ denotes a normalizing constant. The second line is valid because in a polytree, $e_B^+$ and $e_B^-$ are independent given $B$. The factors relating to non-descendants and descendants are denoted $\pi(B)$ and $\lambda(B)$, as shown in the third line of the equation. These factors are called the Pi and Lambda functions, respectively.

The Pi function $\pi(B)$ and Lambda function $\lambda(B)$ can be written as follows:

$$\pi(B) = \sum_{\mathbf{A}} P(B \mid \mathbf{A}) \prod_i \pi_B(A_i) \qquad (3)$$

and

$$\lambda(B) = \prod_j \lambda_{C_j}(B) \qquad (4)$$

where $\pi_B(A_i)$ and $\lambda_{C_j}(B)$ denote a Pi message from the parent $A_i$ to $B$ and Lambda message from the child $C_j$ to $B$, respectively. These messages can be written as follows:

$$\pi_{C_j}(B) = \alpha \left[ \prod_{k \neq j} \lambda_{C_k}(B) \right] \pi(B) \qquad (5)$$

and

$$\lambda_B(A_i) = \sum_B \lambda(B) \sum_{A_k : k \neq i} P(B \mid \mathbf{A}) \prod_{k \neq i} \pi_B(A_k) \qquad (6)$$

Note that the Lambda function $\lambda(B)$ is similar to the Pi message $\pi_{C_j}(B)$ except that the Pi message includes a factor of $\pi(B)$, includes Lambda messages all children of the parent except the target child node $j$, and includes a normalizing constant $\alpha$. A similar relationship exists between the Pi function $\pi(B)$ and the Lambda message $\lambda_B(A_i)$. The Lambda message multiplies by $\lambda(B)$ and includes Pi messages only from parent nodes other than the target parent node $i$.

The MP algorithm is given as follows.

- *Initialization:* For any evidence node $B=b$, set $\pi(B) = \lambda(B)$ to 1 for $B=b$ and 0 for $B \neq b$. For any non-evidence node with no parents, set $\pi(B)$ to the prior distribution for $B$. For any non-evidence node $B$ with no children, set $\lambda(B)$ uniformly equal to 1.

- *Iterate* until no change occurs:
  o For each node $B$, if $B$ has received Pi messages from all its parents, then calculate $\pi(B)$.
  o For each node $B$, if $B$ has received Lambda messages from all its parents, then calculate $\lambda(B)$.
  o For each node $B$, if $\pi(B)$ has been calculated and $B$ has received Lambda messages from all its children except $C$, calculate and send the Pi message from $B$ to $C$.
  o For each node $B$, if $\lambda(B)$ has been calculated and $B$ has received Pi messages from all its parents except $A$, calculate and send the Lambda message from $B$ to $A$.

- *Calculate* $P(B \mid e) = \alpha \pi(B) \lambda(B)$ for each node $B$.

This algorithm finds exact values of all $\pi(B)$ if the network is a polytree. The algorithm can also be applied to BNs containing undirected cycles. It is not guaranteed to converge, but when it converges, it often results in a good approximation to the correct posterior probabilities [Murphy et al, 1999].

*Message Passing Inference for Continuous BNs*

Type 14 in Fig. 1 is a BN in which all nodes are continuous. The MP algorithm can be extended to this case by defining Pi/Lambda messages as follows [Sun, 2007]. These messages can be computed exactly for linear Gaussian networks. For general Gaussian networks, the messages can be approximated using the unscented transformation [Uhlmann, 1995] to project mean and covariance estimates through nonlinear transformations.

The Pi and Lambda functions for a continuous node X in a Gaussian network can be written as follows:

$$\pi(X) = \int_{\mathbf{U}} P(X \mid \mathbf{U}) \prod_i \pi_X(U_i) \, d\mathbf{U} \qquad (7)$$

and

$$\lambda(X) = \prod_j \lambda_{Y_j}(X) \qquad (8)$$

where $d\mathbf{U}$ is the m-dimensional differential with $\mathbf{U} = \{U_1, U_2, ..., U_m\}$, $\pi_X(U_i)$ denotes the Pi message from the continuous parent $U_i$ to $X$, and $\lambda_{Y_j}(X)$ denotes the Lambda message from the continuous child $Y_j$ to $X$. These Pi and Lambda messages can be written as follows:

$$\pi_{Y_j}(X) = \alpha \left[ \prod_{k \neq j} \lambda_{Y_j}(X) \right] \pi(X) \qquad (9)$$

and

$$\lambda_X(U_j) = \int_X \lambda(X) \int_{\overline{\mathbf{U}}} P(X \mid \mathbf{U}) \prod_{k \neq j} \pi_X(U_k) \, d\overline{\mathbf{U}} \, dX \quad (10)$$

where $d\overline{\mathbf{U}}$ is the (m - 1)-dimensional differential with $\overline{\mathbf{U}} = \{s \mid s \in \mathbf{U} \text{ and } s \neq U_j\}$.

### Message Passing Inference for Hybrid BNs

For a conditional hybrid BN (i.e., Types 1, 2, 3, 11, 14, and 17 in Fig. 1), the Pi/Lambda functions and the Pi/Lambda messages can be extended as follows [Sun, 2007].

The Pi function for a discrete node B (i.e., Types 1, 2, and 3) is given by Equation 3. The Pi function for the continuous node X of Type 17 is given as follows:

$$\pi(X) = \sum_{\mathbf{A}} \int_{\mathbf{U}} P(X \mid \mathbf{A}, \mathbf{U}) \prod_i \pi_X(A_i) \prod_j \pi_X(U_j) \, d\mathbf{U} \quad (11)$$

where $d\mathbf{U}$ is the m-dimensional differential with $\mathbf{U} = \{U_1, U_2, ..., U_m\}$, $\pi_X(A_i)$ denotes the Pi message from the discrete parent $A_i$ to $X$ and $\pi_X(U_j)$ denotes the Pi message from the continuous parent $U_j$ to $X$. Derivation of Pi functions for Types 11 and 14 is straightforward by use of Equation 11.

The Lambda function for Types 11, 14, and 17 is given by Equation 8. The Lambda function for the discrete node *B* of Type 3 is given as follows:

$$\lambda(B) = \prod_i \lambda_{C_i}(B) \prod_j \lambda_{Y_j}(B) \quad (12)$$

where $\lambda_{C_i}(B)$ denotes the Lambda message from the discrete child node $C_i$ to $B$ and $\lambda_{Y_j}(B)$ denotes the Lambda message from the continuous child node $Y_j$ to *B*. Derivation of Lambda functions for Types 1 and 2 is straightforward by use of Equation 12.

The Pi message for Types 11, 14, and 17 is given by Equation 9. The Pi message for Type 3 is given as follows:

$$\pi_{Y_j}(B) = \alpha \left[ \prod_i \lambda_{C_i}(B) \prod_{k \neq j} \lambda_{Y_k}(B) \right] \pi(B) \quad (13)$$

The Lambda message for Types 1, 2, and 3 is given by Equation 6. The Lambda message for the continuous node *X* of Type 17 can be written as follows:

$$\lambda_X(U_j) = \int_X \lambda(X) \sum_{\mathbf{A}} \int_{\overline{\mathbf{U}}} P(X \mid \mathbf{A}, \mathbf{U}) \prod_i \pi_X(A_i) \prod_{k \neq j} \pi_X(U_k) \, d\overline{\mathbf{U}} dX \quad (14)$$

and

$$\lambda_X(A_i = a) = \int_X \lambda(X) \sum_{\overline{\mathbf{A}}} \int_{\mathbf{U}} P(X \mid A_i = a, \overline{\mathbf{A}}, \mathbf{U}) \prod_{k \neq i} \pi_X(A_k) \prod_j \pi_X(U_j) \, d\mathbf{U} dX \quad (15)$$

where $d\overline{\mathbf{U}}$ is the (m - 1)-dimensional differential with $\overline{\mathbf{U}} = \{s \mid s \in \mathbf{U} \text{ and } s \neq U_j\}$, $\overline{\mathbf{A}} = \{s \mid s \in \mathbf{A} \text{ and } s \neq A_i\}$, and $a$ denotes a state of $A_i$. The first equation is for the message to the continuous parent $U_j$ and the second equation is for the message to the discrete parent $A_i$. These apply respectively to Types 11 and 14.

### B. Gaussian Mixture Reduction

Gaussian mixture reduction (GMR) approximates an M-component GMM with a reduced number N<M of components. Several methods for GMR have been proposed, *e.g.*, [Salmond, 1990][West, 1993][Williams, 2003][Williams & Maybeck, 2003][Schrempf et al., 2005][Runnalls, 2007].

A straightforward method for performing GMR is the following:

(1) Find the two closest components in a GMM according to a distance criterion.

(2) Merge the two selected components into one component.

(3) Update to a GMM with one fewer component.

(4) Repeat steps 1-3 until a stopping criterion is reached (e.g., a predefined number of components, a predefined precision, etc.).

As a distance criterion, Runnalls [2007] proposed the Kullback-Leibler (KL) divergence [Kullback & Leibler, 1951]. The distance criterion using the KL divergence is written as follows [Runnalls, 2007].

$$d\left((w_i, \mu_i, \sigma_i), (w_j, \mu_j, \sigma_j)\right) \quad (16)$$
$$= \frac{\left[(w_i + w_j) \log \det(\sigma_{ij}) - (w_i) \log \det(\sigma_i) - (w_j) \log \det(\sigma_j)\right]}{2}$$

where *i* and *j* denote the *i-th* and *j-th* component of a GMM, respectively, and $w_k$, $\mu_k$, $\sigma_k$ are the weight, mean, and covariance of the *k-th* component, respectively.

Recently, a more efficient algorithm using constraint optimization was proposed [Chen et al., 2010][Chang & Sun, 2010].

## III. EXTENDED HYBRID MESSAGE PASSING ALGORITHM

The previous sections introduced message passing inference and Gaussian mixture reduction. This section combines these methods into an extended hybrid message passing algorithm for CG BNs.

The GMR operation is denoted as a function, $\tau(gmm, max\_nc)$ that applies Equation 16, where $gmm$ is a Gaussian mixture model and $max\_nc$ is the maximum number of allowable mixture components.

To specify the algorithm, we need to define where in the inference process GMR will be applied. For this, the Pi/Lambda functions for $X$ and Pi/Lambda messages $U \rightarrow X$ in Type 14 and $\{A, U\} \rightarrow X$ in Type 17 are chosen. Hence, the function $\tau(gmm, max\_nc)$ is applied to Equations 7, 8, 9, 10, 11, 14, and 15. For the extended algorithm, these equations become:

$$\pi(X) = \tau\left(\int_{\mathbf{U}} P(X \mid \mathbf{U}) \prod_i \pi_X(U_i) \, d\mathbf{U}, M\right) \quad (17)$$

$$\pi(X) = \tau\left(\sum_{\mathbf{A}} \int_{\mathbf{U}} P(X \mid \mathbf{A}, \mathbf{U}) \prod_i \pi_X(A_i) \prod_j \pi_X(U_j) \, d\mathbf{U}, M\right) \quad (18)$$

$$\lambda(X) = \tau\left(\prod_j \lambda_{Y_j}(X), M\right) \quad (19)$$

$$\pi_{Y_j}(X) = \alpha \tau\left(\prod_{k \neq j} \lambda_{Y_j}(X), M\right) \pi(X) \quad (20)$$

$$\lambda_X(U_j) = \int_X \lambda(X) \tau\left(\int_{\bar{\mathbf{U}}} P(X \mid \mathbf{U}) \prod_{k \neq j} \pi_X(U_k) \, d\bar{\mathbf{U}}, M\right) dX \quad (21)$$

$$\lambda_X(U_j) = \int_X \lambda(X) \tau\left(\sum_{\mathbf{A}} \int_{\bar{\mathbf{U}}} P(X \mid \mathbf{A}, \mathbf{U}) \prod_i \pi_X(A_i) \prod_{k \neq j} \pi_X(U_k) \, d\bar{\mathbf{U}}, M\right) dX \quad (22)$$

$$\lambda_X(A_i = a) = \int_X \lambda(X) \tau\left(\sum_{\bar{\mathbf{A}}} \int_{\mathbf{U}} P(X \mid A_i = a, \bar{\mathbf{A}}, \mathbf{U}) \prod_{k \neq i} \pi_X(A_k) \prod_j \pi_X(U_j) \, d\mathbf{U}, M\right) dX \quad (23)$$

where $M = max\_nc$ denotes the maximum allowable number of components.

The above equations are implemented in the following algorithm, called *Hybrid Message Passing Algorithm with Gaussian Mixture Reduction* (HMP-GMR). The following algorithm is an extension of a Hybrid Message Passing algorithm (HMP) from [Sun, 2007] to which we apply GMR. of their anytime property. An initial version of this algorithm was introduced in [Park et al., 2015]. In contrast with the initial version, this is an anytime algorithm that can provide a solution even if it is interrupted before completion.

---

**Algorithm 1**: Hybrid Message Passing (HMP) with Gaussian Mixture Reduction (GMR) Algorithm

**Procedure** HMP-GMR (
    *net*,   // a Hybrid BN
    *max_time*,   // a maximum execution time
    *max_iteration*, // a maximum number of iterations
    *max_nc*,   // a maximum allowable number of components
    *max_prcs*   // a maximum precision
)
1  **for** $i = 1, \ldots$ **until** *max_iteration*
2   **for** $j = 1, \ldots$ **until** the number of nodes in *net*
3    $\pi_j \leftarrow$ ComputeAllPiMsgs($j$, *max_nc*, *max_time*)
4    $\lambda_j \leftarrow$ ComputeAllLambdaMsgs ($j$, *max_nc*, *max_time*)
5    SendPiMsg($j$, *max_nc*, *max_time*)
6    SendLambdaMsg ($j$, *max_nc*, *max_time*)
7   **for** $j = 1, \ldots$ **until** the number of nodes in *net*
8    $bel_{ij} \leftarrow$ compute belief function using $\lambda_j$ and $\pi_j$ (2)
9    $diff_{ij} \leftarrow$ compare distribution difference between $bel_{ij}$ and $bel_{(i-1)j}$
10   $max\_diff \leftarrow$ get maximum difference between $diff_{ij}$ and $max\_diff$
11   **if** $max\_diff < max\_prcs$ **then** break
12 **return** a set of $bel_{ij}$

**Procedure** ComputeAllPiMsgs (
    $j$,     // a current node
    *max_time*, // a maximum execution time
    *max_nc*   // a maximum allowable number of components
)
1 **if** $j$ is discrete **then** do (3)
2 **else if** $j$ is continuous then
3   **if** parent of $j$ is discrete **then** do (11)
4   **if** parent of $j$ is continuous **then** do (17) with $M = max\_nc$
5   **if** parent of $j$ is hybrid **then** do (18) with $M = max\_nc$
6 $exe\_time \leftarrow$ get a current execution time
7 **if** $max\_time < exe\_time$ do not update $\pi_j$
8 **return** $\pi_j$

**Procedure** ComputeAllLambdaMsgs (
    $j$,     // a current node
    *max_time*, // a maximum execution time
    *max_nc*   // a maximum allowable number of components
)
1 **if** $j$ is discrete **then** do (4)
2 **else if** $j$ is continuous **then** do (19) with $M = max\_nc$
3 $exe\_time \leftarrow$ get a current execution time
4 **if** $max\_time < exe\_time$ do not update $\lambda_j$
5 **return** $\lambda_j$

**Procedure** SendPiMsg(
    $j$,     // a current node
    *max_time*, // a maximum execution time
    *max_nc*   // a maximum allowable number of components
)
1 **for** $k = 1, \ldots$ **until** number of children of $j$
2   **if** $j$ is discrete **then** do (5) for $k$
3   **else if** $j$ is continuous **then** do (20) for k with $M = max\_nc$
4 $exe\_time \leftarrow$ get a current execution time
5 **if** $max\_time < exe\_time$ do not update the Pi message from one of (5) and (20)

**Procedure** SendLambdaMsg (
    $j$,     // a current node
    *max_time*, // a maximum execution time
    *max_nc*   // a maximum allowable number of components
)
1 **for** $k = 1, \ldots$ **until** number of parents of $j$
2   **if** $j$ is discrete **then** do (6) for $k$
3   **else if** $j$ is continuous **then**
4     **if** parent of $j$ is discrete **then** do (15)
5     **if** parent of $j$ is continuous **then** do (21) with $M = max\_nc$
6     **if** parent of $j$ is hybrid **then** do (22), (23) with $M = max\_nc$
7 $exe\_time \leftarrow$ get a current execution time
8 **if** $max\_time < exe\_time$ do not update the Lambda message from one of (6), (15), (21), (22), and (23)

HMP-GMR has five inputs. The first input, *net*, is the Hybrid BN with specified evidence nodes and their values. The second input, *max_time*, is the maximum execution time used to control how long this algorithm runs by comparing to a current execution time, *exe_time*, representing a period from the algorithm starting time to the current time. The third input, *max_iteration*, is the maximum number of iterations allowed, where an iteration is one round in which all nodes in the BN perform their operations. The fourth input, *max_nc,* is the maximum number of Gaussian components that may be output by the GMR function τ. The fifth input, *max_prcs*, is a threshold on the distance between posterior distributions of nodes in the current and previous iterations. The algorithm terminates when the distance is lower than the threshold. HMP-GMR outputs approximate posterior distributions of all nodes.

Given these inputs, the algorithm proceeds as follows: (1) The algorithm iterates message passing from 1 to the maximum number of iterations or until it is interrupted due to exceeding the time limit. (2) The algorithm cycles through all nodes in the BN. (3) For the *j-th* node, all Pi messages from its parents are computed to calculate the Pi value $\pi_j$. If the RV is discrete, Equation 3 is used, while if it is continuous and has only discrete, only continuous, or hybrid parents, Equation 11, 17, or 18, is used, respectively. (4) All Lambda messages from children of the *j-th* node are computed to calculate the Lambda value $\lambda_j$. If the RV is discrete, Equation 4 is used, while if it is continuous Equation 19 is used. (5) A Pi message is sent from the *j*-th node to its children. If the node is discrete, Equation 5 is used, while if it is continuous Equation 20 is used. (6) A Lambda message is sent from the *j-th* node to its parents. If the node is discrete, Equation 6 is used, while if it is continuous and has only discrete, only continuous, or hybrid parents, Equation 15, 21, or 22 (for continuous parent) / 23 (for discrete parent) is used, respectively. For each of these functions in Lines 3, 4, 5, and 6, if the current execution time exceeds the maximum execution time, the result from the function is not updated and the for-loop in Line 2 of the HMP-GMR procedure is stopped. After all nodes have passed their messages (i.e., Line 2 ~ 6), (7) the belief function is computed for all nodes. (8) The Lambda and Pi values are multiplied and normalized for all nodes to calculate the belief function $bel_{ij}$. (9) The difference $diff_{ij}$ between the current and previous beliefs are computed for all nodes. (10) The maximum difference *max_diff* between current and previous belief is selected. (11) If the maximum difference *max_diff* is less than the maximum precision *max_prcs*, the iteration of the message passing is stopped. (12) Upon stopping, the algorithm outputs approximate posterior marginal distributions for all nodes.

There are three exit points from the iteration: (1) when the iteration reaches the maximum number of allowable iterations, (2) when the maximum difference is less than the maximum precision, and (3) when the current execution time for the algorithm exceeds the maximum execution time.

IV. Optimizing the Settings of HMP-GMR

In some situations, the Hybrid Message Passing with Gaussian Mixture Reduction (HMP-GMR) algorithm performs better than other algorithms. For example, although in theory a sampling algorithm can be made as accurate as desired, for a given problem, HMP-GMR may have higher accuracy for a given limit on computation time. However, HMP-GMR requires initial settings before it executes. The performance of HMP-GMR depends on these initial settings. More specifically, HMP-GMR requires that the maximum allowable number of components *max_nc* and the maximum number of allowable iterations *max_iteration* are specified as inputs. If the maximum allowable number of components is too small, accuracy may be too low; but if it is too large, execution time may be unacceptably long. Also, the maximum number of allowable iterations can influence accuracy and execution time. The number of components required to achieve a given accuracy depends on the network topology, the placement of continuous and discrete nodes, and the conditional distributions. As noted above, when the BN contains loops, the HMP-GMR algorithm may not converge. Thus, in some problems, HMP-GMR may spend many iterations without a significant improvement in accuracy.

Therefore, there is a need to trade off accuracy against execution time depending on the maximum allowable number of components and the maximum number of iterations. Different applications pose different requirements on execution time. It is assumed that the maximum allowable execution time for inference is an input parameter that is specified before the inference algorithm runs. Therefore, the optimization problem is defined as attaining the best achievable accuracy for a given constraint on execution time, by varying the maximum allowable number of components and the maximum allowable number of iterations.

Finding an exact optimum would be infeasible in the general case. Therefore, this section presents a Monte Carlo method to find approximately optimal values for a specific conditional Gaussian Bayesian network. The algorithm is appropriate for problems in which a given HBN is specified *a priori*, and inference on the HBN will be performed repeatedly in a time-restricted setting with limits on execution time for inference. The optimization can be performed offline as a preprocessing step to find good initial settings for performing HMP-GMR inference at run time. For example, a real-time threat detection system might use a CG HBN to process sensor inputs automatically and issue alarms when suspicious combinations of sensor readings occur. Because the system runs in real time, fast execution is essential. At design time, an offline optimization can be run using the algorithm presented here to determine the best settings for the maximum number of components and the maximum number of iterations.

An optimization problem for this situation can be formulated as shown Equation 24. In this setting, we assume that a specific HBN is given.

$$\text{Min}_{nc \in \mathbf{NC}, it \in \mathbf{IT}} f(nc, it) \quad (24)$$

subject to: *t* < *max_time*

where $\mathbf{NC}$ means a set of the candidate maximum allowable numbers of components $\{nc_1, nc_2, ..., nc_n\}$, $\mathbf{IT}$ means a set of the candidate maximum iteration numbers $\{it_1, it_2, ..., it_m\}$, $f(.)$ is an objective function measuring error of HMP-GMR, *t* means the current execution time for inference of HMP-GMR, and *max_time* means the maximum execution time. We call this as HMP-GMR with Optimal Settings (HMP-GMR-OS), which finds the values (*nc* and *it*) that achieve the best accuracy under a given time restriction. Equation 25 shows the objective function *f*(.) of HMP-GMR-OS.

$$f(nc, it) = \frac{\sum_{e \in E} err(s(net, e), h(net, e, max\_time, nc, it))}{|E|} \quad (25)$$

where $E$ means a set of the candidate evidences $\{e_1, e_2, ..., e_l\}$, which are randomly selected, for a Bayesian network *net*, *err*(.) means a function resulting in an error between a near-correct inference result from sampling and an inference result from HMP-GMR, *s*(.) means a sampling inference algorithm used for exact inference, *h*(.) means the HMP-GMR algorithm with a maximum execution time *max_time*, a candidate maximum allowable number of components *nc*, and a candidate maximum iteration number *it*.

The Monte Carlo method for HMP-GMR-OS is called an HMP-GMR-OS algorithm (Algorithm 2), which finds the best values for the two decision variables given a Hybrid BN, a maximum execution time, a number of samples, an upper limit on the maximum number of iterations, and an upper limit on the maximum allowable number of components.

**Algorithm 2**: HMP-GMR with Optimal Settings (HMP-GMR-OS) Algorithm
**Procedure** HMP-GMR-OS (
    *net*,      // a Hybrid BN
    *max_time*,      // a maximum execution time
    *num_samples*,      // a number of samples
    *ul_max_it*,      // an upper limit on the maximum number of iterations
    *ul_max_nc*      // an upper limit on the maximum allowable number of components
)
1  **for** *i* = 1, … **until** *num_samples*
2    $e_i$ ← generate randomly a set of evidence values from *net*
3    $s_i$ ← inference using a sampling algorithm with *net* and $e_i$
4    **for** *j* = 1, … **until** *ul_max_nc*
5      **for** *k* = 1, … **until** *ul_max_it*
6        $h_{jk}$ ← inference using HMP-GMR with *net*, $e_i$, *max_time*, *j*, and *k*
7        $r_{ijk}$ ← calculate an inference error value between $s_i$ and $h_{jk}$
8        $r_{jk}$ ← add $r_{ijk}$ to a set of inference error values $r_{jk}$ for *j* and *k*
9  **for** *j* = 1, … **until** *ul_max_nc*
10    **for** *k* = 1, … **until** *ul_max_it*
11      $avg\_r_{jk}$ ← calculate an average inference error for $r_{jk}$
12  [*nc*, *it*] ← select best values for *nc* and *it* in (25) using $avg\_r_{jk}$
13  **return** [*nc*, *it*]

The HMP-GMR-OS algorithm has five inputs. The first input *net* is a Hybrid BN. The second input *max_time* is the maximum execution time for inference of HMP-GMR. The third input *num_samples* is the number indicating how many times the simulation should be repeated. The fourth input *ul_max_it* is the number of maximum iterations used for inference of HMP-GMR. The fifth input *ul_max_nc* is the number indicating how many the number of maximum allowable number of components will be investigated.

Given these inputs, the algorithm proceeds as follows: (1) The algorithm simulates the given number of samples. (2) The algorithm randomly selects some evidence nodes from the Hybrid BN *net*. Also, it randomly selects a reasonable evidence value for each evidence node (i.e., a highly unlikely value is not used for the evidence value) and provides a *i*-th set of evidence values $e_i$. (3) The set of the evidence values are used for inference of a sampling algorithm by which nearly correct results of inference $s_i$ (i.e., posterior distributions) are found. (4) The maximum allowable number of components, denoted by *j*, is varied from 1 to the upper limit on the maximum allowable number of components *ul_max_nc*. (5) The maximum number of iterations, denoted by *k*, is varied from 1 to the upper limit on the maximum number of iterations *ul_max_it*. (6) This algorithm uses the HMP-GMR algorithm with the Hybrid BN *net*, the set of the evidence vlaues $e_i$, the maximum execution time *max_time*, the maximum allowable number of components *j*, and the maximum number of iterations *k*. Then, the HMP-GMR algorithm provides the results $h_{jk}$ (i.e., posterior distributions). (7) An inference error value $r_{ijk}$ between the nearly correct results $s_i$ and the HMP-GMR's result $h_{jk}$ is computed by using a distance function (e.g., KL-divergence [Kullback & Leibler, 1951]). (8) The inference error value $r_{ijk}$ at *i*-th sample for *j* and *k* is stored at a set of inference error values $r_{jk}$ for *j* and *k*. After simulating all samples, for all *j* (9) and *k* (10), (11) an average inference error $avg\_r_{jk}$ is calculated using the set of the inference error values $r_{jk}$. (12) A best maximum allowable number of components *nc* and a best maximum number of iterations *it* are selected by finding a minimum average inference error from $avg\_r_{jk}$. (13) The algorithm outputs the best values *nc* and *it*.

In summary, the HMP-GMR-OS algorithm is a preprocessing algorithm finding the optimal settings for HMP-GMR given a HBN to improve accuracy before HMP-GMR for the HBN executes for a practical situation.

V. EXPERIMENT

This section presents experiments to evaluate the performance of the HMP-GMR algorithm and the Optimal GMR algorithm. For evaluation of the HMP-GMR algorithm, Park et al. [2015] presented simple experiments to demonstrate scalability and efficiency using two hybrid BNs. Here, more extensive experiments are performed on four BNs. These four BNs would be representative BNs in terms of various numbers of discrete parent nodes and various numbers of loopy structures in a given network.

Fig. 2 shows two illustrative Conditional Gaussian (CG) BNs (i.e., 1 and 2) containing a discrete node *A* with 4 states, a continuous node $X_j$, and another continuous node $Y_j$. These two BNs differ in the links between $Y_j$ and $Y_{j+1}$. The first, shown on the left in Fig. 2, has no undirected cycles involving only continuous nodes, while the second, shown on the right in Fig. 2, has undirected cycles among. For example, between $X_1$ and $Y_2$, there are two paths: $X_1 \rightarrow X_2 \rightarrow Y_2$ and $X_1 \rightarrow Y_1 \rightarrow Y_2$.

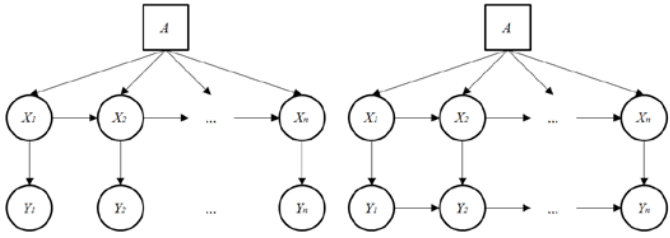

Fig. 2. Conditional Gaussian BN 1 in the left figure and Conditional Gaussian BN 2 in the right figure

Fig. 3 shows two additional cases in which the BNs contain a large number of discrete nodes. Each discrete node $A_i$ has four states and the continuous nodes $X_i$ and $Y_j$ are conditional Gaussians. Again, the difference between the two BNs is that the third has no undirected cycles involving only continuous nodes, while the fourth has loopy structure for the continuous nodes.

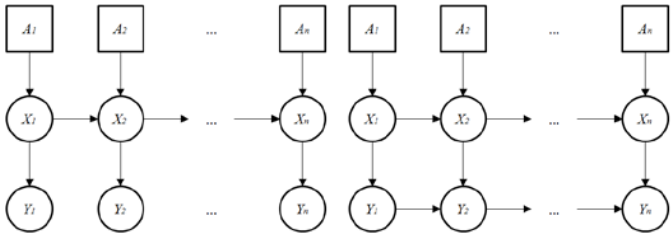

Fig. 3. Conditional Gaussian BN 3 in the left figure and Conditional Gaussian BN 4 in the right figure

The experiments examined both CLG and CNG BNs with these four CG BNs. The size of the four CG BNs was varied by adjusting $n$. Some of the leaf continuous nodes $\{Y_1, \ldots, Y_n\}$ were randomly selected as evidence nodes. All other nodes were unobserved.

Table 2 shows characteristics of these four CG BNs. In these BNs, the discrete nodes have four states. Therefore, in CG BNs 1 and 2, there are four discrete states for discrete node $A$, while in CG BNs 3 and 4, there are $4^n$ total configurations of the discrete states. This is $4^7 = 16384$ configurations when $n = 7$. When $n = 7$, CG BNs 1 and 4 contain 21 cycles, while CG BN 2 contains 501 cycles[2]. CG BN 3 contains no cycles.

|  | CG BN 1 | CG BN 2 | CG BN 3 | CG BN 4 |
|---|---|---|---|---|
| Cycles | 21 | 501 | 0 | 21 |
| Combinations of discrete states | 4 | 4 | 16384 | 16384 |

Table 2. Characteristics of four CG BNs with $n = 7$

The following factors were varied in the experiment: (1) Type of hybrid BN (i.e., BN 1, 2, 3, or 4; CLG or CNG BN), (2) type of inference algorithm (i.e., Hybrid Junction Tree (Hybrid-JT) [Lauritzen, 1992][Lauritzen & Jensen, 2001], original Hybrid MP (HMP) [Sun, 2007], Hybrid MP with Gaussian Mixture Reduction (HMP-GMR) [Park et al., 2015], or Likelihood Weighting (LW) sampling [Fung & Chang, 1989]), (3) number of repeated structures $n$, (4) algorithm characteristics (*i.e.*, the number of GMM components allowed, the number of allowable message passing iterations, and the maximum precision for the message passing algorithms). The dependent variables were accuracy of result and execution time.

---

[2] Cycles were derived by a cycle finding algorithm [Johnson, 1975].

For all experiments, the convergence criterion for HMP and HMP-GMR was $max\_prs = 10^{-3}$.

Using these settings, we conducted three experiments: (1) A comparison between HMP, HMP-GMR, and Hybrid-JT investigated scalability of HMP-GMR to complex networks (i.e., larger values of $n$), (2) the HMP-GMR itself was evaluated for posterior distribution accuracy and execution time, and (3) optimal settings derived from the HMP-GMR-OS algorithm were evaluated by inference accuracy. The experiments were run on a 3.40GHz Intel Core i7-3770 processor. The algorithms were implemented in the Java programming language. In a Java code for HMP-GMR, there was another exit point from the iteration of the HMP-GMR algorithm in Section 3. In some cases, a computation between Lambda values and Pi values in the HMP-GMR algorithm could not be computed due to numeric underflow. This happened when the HMP-GMR algorithm diverged. When this occurred, the HMP-GMR algorithm stopped and provided its current solution.

### A. Scalability of HMP-GMR

The first experiment examined improvement in scalability of HMP-GMR over HMP and Hybrid-JT.

The initial setting of this experiment consists of (1) maximum of 4 components output by GMR and (2) 100 iteration limit for each of HMP and HMP-GMR. Eight CG BNs (i.e., conditional linear/nonlinear cases for CG BNs 1, 2, 3, and 4) were run with HMP, HMP-GMR, and Hybrid-JT using the following inputs and outputs. The input value of $n$ for both BNs was varied from 1 to 10. The output value is the execution time.

Fig. 4 and 5 show the results of this experiment summarizing the execution times for the CLG case as the number of nodes $n$ is varied. The X axis for each figure denotes the number of nodes $n$. The Y axis for each figure denotes the execution time in milliseconds. The solid line denotes the HMP-GMR results. The dashed line denotes the HMP results. The dotted line denotes the Hybrid-JT results.

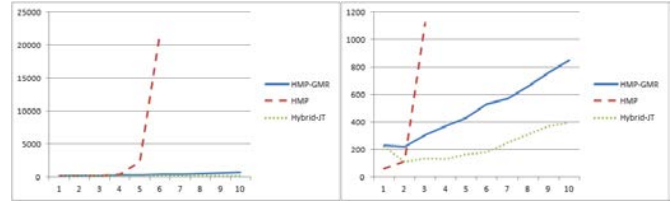

Fig. 4. Execution Times over $n$ on CG BNs 1 and 2

Results for CG BNs 1 and 2 show a similar pattern. The HMP algorithm with no GMR exceeded the time limit at $n = 7$ and $n = 4$, respectively. Execution times for HMP-GMR were higher than those for Hybrid-JT in both cases. The increase in execution time for both HMP-GMR and Hybrid-JT was linear in $n$.

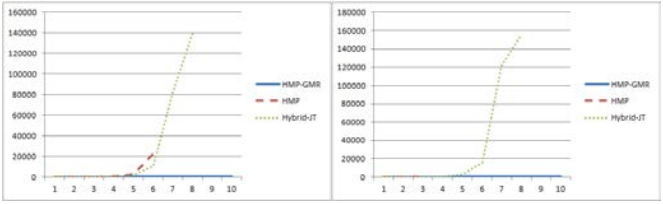

Fig. 5. Execution Times over *n* on CG BNs 3 and 4

In Fig. 5, results from HMP and Hybrid-JT show exponential growth in execution time, while execution time of HMP-GMR increased linearly. For HMP, the execution time limit for CG BNs 3 and 4 was exceeded at $n = 7$ and $n = 4$, respectively. For Hybrid-JT, the execution time limit for CG BNs 3 and 4 was exceeded at $n = 9$.

Results for the CNG networks showed similar patterns, and are not shown here for brevity. These experiments showed that HMP-GMR is scalable to large BNs for both linear and nonlinear CG networks. However, scalability alone is not sufficient. Accuracy and good operational performance are also essential.

*B. Accuracy and Efficiency of HMP-GMR*

In this experiment, we investigated the accuracy and convergence of HMP-GMR for the four CLG BNs. To evaluate accuracy of HMP-GMR, exact inference results using Hybrid-JT inference were used. Some of the runs using Hybrid-JT stopped because of the exponential growth of components, so Hybrid-JT produced posterior distributions only for $n \leq 7$. For this reason, this experiment used $n = 7$ for the four CLG BNs. Accuracy was measured by KL-divergence [Kullback & Leibler, 1951] (lower values mean better accuracy). We calculated the KL-divergence between exact and approximate results for each unobserved node, and summed them (henceforth, we use KL-divergences to mean the sum of KL-divergences over unobserved nodes). The number of runs in the experiment was 100. The maximum allowable number of components was $nc = 2$. The maximum number of iterations was $it = 10000$. The maximum execution time was $max\_time = 200000$ millisecond (ms). In the experiment, there were three exit points: (1) When the algorithm converged, (2) when the time limit was exceeded, and (3) when the algorithm diverged. When the algorithm did not converge, the algorithm halted and provided its current solution.

Fig. 6 shows percentages for each CLG BN for which the algorithm converged, diverged, and ran out of time. For CLG BN 1, the algorithm converged in 97% of the runs and ran out of time in 3% of the cases (execution time > 200000 ms). For CLG BN 2, the algorithm converged in 52% of the runs, ran out of time in 3% of the runs, and diverged in 45% of the runs. For CLG BN 3, the algorithm converged in 100% of the runs. For CLG BN 4, the algorithm converged in only 31% of the runs and ran out of time in 69% of the runs.

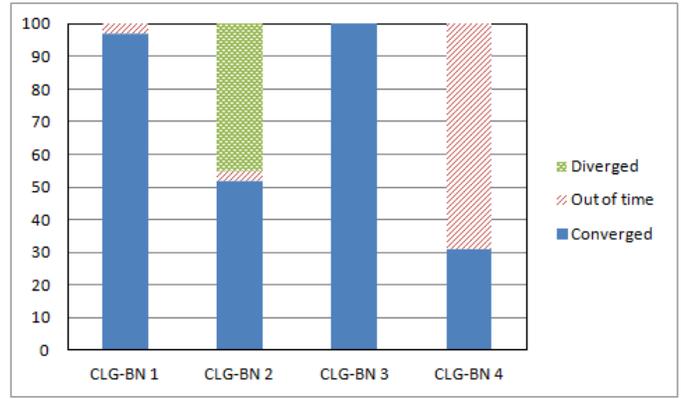

Fig. 6. Percentages for each model case when the algorithm converged, diverged, and ran out of time

We observed that the large number of cycles (CLG BN 2) could cause many situations in which the algorithm did not converge (i.e., diverged or ran out of time). Also, the algorithm for the cases with no cycles (CLG BN 3) always converged. For the cases in which the algorithm ran out of time, if more time had been allowed it might have converged, diverged, or failed to either converge or diverge (i.e., oscillated). In some cases for CLG BNs 1, 2, and 4, the algorithm oscillated until reaching the maximum execution time and halting. When there were many cycles and many discrete states (i.e., for CLG BN 4), the algorithm often ran out of time.

|  |  | CLG BN 1 | CLG BN 2 | CLG BN 3 | CLG BN 4 |
|---|---|---|---|---|---|
| Converged | avg. KL-divergence | 0.0001 (0.0001) | 1.0352 (1.4604) | 2.2469 (1.452) | 3.6409 (3.1155) |
|  | avg. Time | 1514 (363.53) | 16547 (24113) | 2163.7 (415.6) | 9584 (9739.4) |
| Diverged | avg. KL-divergence | - | 106.3524 (16.0374) | - | - |
|  | avg. Time | - | 9677.7 (5703.3) | - | - |
| Out of time | avg. KL-divergence | 0.0528 (0.0545) | 3.0795 (2.3361) | - | 82.0464 (121.4148) |

Table 3. Average accuracies and average execution times for the four CLG BNs

Table 3 shows averages (avg.) for KL-divergence over runs and average execution times for three cases (converged, diverged, and ran out of time) on the four CLG BNs (numbers in parentheses are standard deviations). Fig. 7 shows the accuracy results from this experiment, when the algorithm converged. In Fig. 7, the four lanes denote the four CLG BNs. The vertical axis on the upper chart denotes KL-divergence values. For case of convergence, the averages for KL-divergence over runs of the experiment were 0.0001, 1.04, 2.25, and 3.64 for CLG BNs 1, 2, 3, and 4, respectively. Again for the case of convergence, the average execution times were 1514, 16547, 2164, and 9584 for CLG BNs 1, 2, 3, and 4, respectively. For the case of convergence, because of the large number (501) of cycles in CLG BN 2, the inference algorithm required more time (avg. 16547 ms) to converge than others. Also, the algorithm for CLG BN 4 spent more time (avg. 9584 ms) in comparison with the case (avg. 2164 ms) of CLG BN 3, because of the large number (21) of cycles in CLG BN 4. In the case of divergence, the algorithm for CLG BN 2 halted with numeric underflow in the Lambda or Pi value computation. In this case, the algorithm performed with very poor accuracy (avg. 106.35) and had a long execution time (avg. 9678 ms).

Some cases for CLG BNs 1, 2, and 4 stopped because the maximum execution time was exceeded. This never happened in CLG BN 3, which contained no cycles. In addition, accuracies for CLG BNs 1 and 2 were better than those for CLG BN 4.

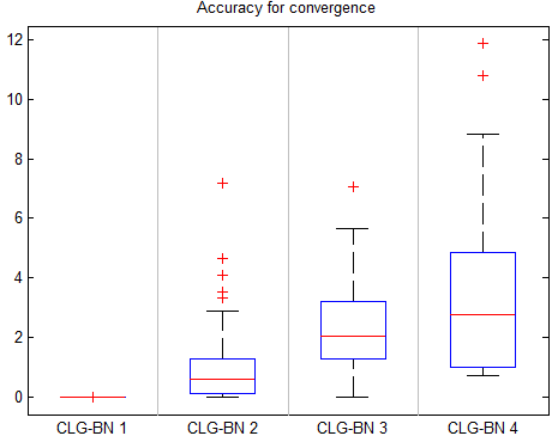

Fig. 7. Accuracies for convergence on the four CLG BNs

The four sets of results depicted in Table 3 illustrated how network topology influences accuracy and execution time. In this experiment, we used arbitrary settings (i.e., $nc = 2$ and $it = 10000$) for HMP-GMR. In the next section, we investigate whether the performance of HMP-GMR can be improved by optimizing these settings.

### C. Optimal Settings for HMP-GMR

HMP-GMR requires initial settings (i.e., $nc$ and $it$), which influence accuracy and execution time. To find good initial settings, the HMP-GMR-OS algorithm was introduced in Section 4. With this algorithm, we use the following experiment setting: (1) the Hybrid BNs (i.e., CLG BNs 1, 2, 3, and 4 with $n = 7$), (2) the maximum execution time (i.e., $max\_time = 3000$ ms), (3) the number of samples (i.e., $num\_samples = 50$), (4) the upper limit on the maximum allowable number of components (i.e., $ul\_max\_nc = 10$), (5) the upper limit on the maximum number of iterations (i.e., $ul\_num\_it = 10$), and (6) the Hybrid-JT algorithm to obtain correct inference results.

Fig. 8 shows the results from this experiment obtained by the HMP-GMR-OS algorithm. Results for CLG BNs 1, 2, 3 and 4 are shown at the top-left, top-right, bottom-left, and bottom-right, respectively. The vertical axes on the four charts denote average KL-divergence values. The bottom-left axis for each chart denotes the maximum number of iterations, while the bottom-right axis denotes the maximum allowable number of components.

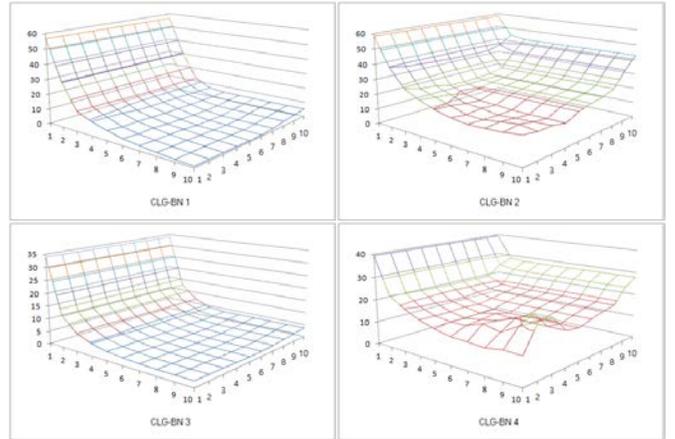

Fig. 8. Accuracy as a function of maximum allowable numbers of components and maximum iterations for four CLG BNs

Table 4 shows minimum averages for KL-divergences in Fig. 8 and best values for $nc$ and $it$ (numbers in parentheses are standard deviations). For example, for CLG BN 1, the minimum average for KL-divergence was 0.78 and its standard deviation was 3.37 at $nc = 5$ and $it = 10$. For CLG BN 4, the minimum average for KL-divergence was 11.37 and its standard deviation was 8.44 at $nc = 1$ and $it = 8$.

|  | CLG BN 1 | CLG BN 2 | CLG BN 3 | CLG BN 4 |
|---|---|---|---|---|
| Minimum average | 0.7771 (3.3748) | 15.1868 (20.999) | 1.9617 (1.72) | 11.3698 (8.4352) |
| Best $nc$ | 5 | 4 | 6 | 1 |
| Best $it$ | 10 | 8 | 10 | 8 |

Table 4. Minimum averages for KL-divergences and best values for $nc$ and $it$

For CLG BNs 1 and 3, the optimal number of iterations was the upper limit of 10. A better value might have been found if a larger number of iterations had been investigated. For CLG BNs 2 and 4, the best value for the maximum number of iterations was smaller than 10. Note that better results might be obtained, if the number of samples was increased and/or the ranges of $nc$ and/or $it$ were expanded.

From this experiment, we observed that good accuracy can be achieved with a small number of components. Although the best values found by our experiment for $nc$ were 5, 4, 6, and 1 for CLG BNs 1, 2, 3, and 4, respectively, the accuracy was not much better than using a single component. For example, the average of KL-divergence for CLG BN 1 was 0.78 at $nc = 5$ and $it = 10$, while the average of KL-divergence for CLG BN 1 was 1.09 at $nc = 1$ and $it = 10$. To check whether the difference was statistically significant, paired t-tests were performed at the 5% significance level. Table 5 shows confidence intervals from the paired t-tests. From these tests, we observed that the difference between the setting found by HMP-GMR-OS and the case with $nc = 1$ and $it = 10$ was not statistically significant for any of the CLG BNs.

|  | CLG BN 1 | CLG BN 2 | CLG BN 3 | CLG BN 4 |
|---|---|---|---|---|
| Confidence Interval | -1.7623, 1.1444 | -12.7994, 5.5235 | -1.1426, 0.1924 | -4.4534, 2.7875 |

Table 5. Confidence intervals from paired t-tests between the optimal settings and the default settings with $nc = 1$ and $it = 10$

This result suggests choosing $nc = 1$ as a default setting for HMP-GMR. This default setting for HMP-GMR was evaluated for accuracy against the Likelihood Weighting (LW) algorithm. Fig. 9 shows accuracy comparison between HMP-GMR with the default setting $nc = 1$ and LW sampling for the four CLG BNs. We use the following Experiment setting: (1) type of the hybrid BNs (i.e., CLG BNs 1, 2, 3, and 4 with $n = 7$), (2) type of inference algorithm (i.e., HMP-GMR at $nc = 1$ and $it = 10$, and LW sampling), (3) 100 samples, (4) the maximum execution time (i.e., *max_time* = 3000 ms), and (5) the Hybrid-JT algorithm to obtain correct inference results.

Fig. 9 shows results from this experiment. When HMP-GMR didn't converge, it stopped at the maximum execution time and provided its current solution. The vertical axis denotes KL-divergence values. The chart contains eight lanes for four groups (CLG BNs 1, 2, 3, and 4). In the two adjoined lanes for each group, the left lane denotes the HMP-GMR case, while the right lane denotes the LW case. For example, the first lane in Fig. 9 denotes the HMP-GMR case for CLG BN 1 (i.e., HG 1), while the second lane in Fig. 9 denotes the LW case for CLG BN 1 (i.e., LW 1). The execution times for the two cases in each group were set to similar values. That is, the number of samples for LW was controlled to achieve similar execution times as HMP-GMR.

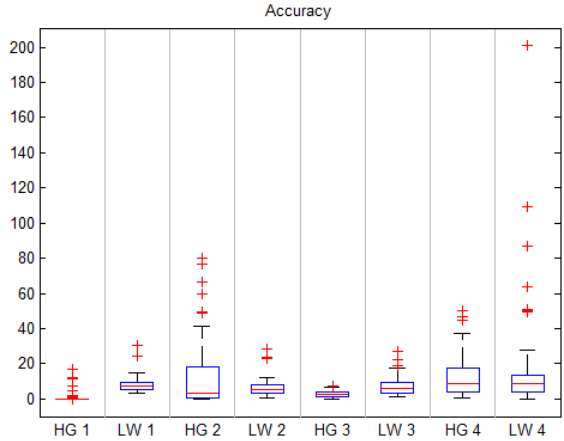

Fig. 9. KL-divergences of HMP-GMR (HG) and LW for four CLG BNs

Table 6 shows averages of KL-divergences for the two algorithms (numbers in parentheses are standard deviations). For example, for CLG BN 1, an average KL-divergence from HMP-GMR was 0.57. For CLG BN 4, an average KL-divergence from LW was 14.29. The fourth row denotes a natural-log ratio between HMP-GMR and LW. In the comparison between HMP-GMR and LW, LW was better than HMP-GMR for CLG BN 2.

|  | CLG BN 1 | CLG BN 2 | CLG BN 3 | CLG BN 4 |
| --- | --- | --- | --- | --- |
| HMP-GMR | 0.5665 (2.4692) | 12.224 (18.0106) | 2.6175 (1.6649) | 11.6846 (10.8626) |
| LW | 7.8211 (4.127) | 6.1218 (5.0357) | 4.7952 (6.865) | 14.2882 (24.955) |
| LN(HMP-GMR / LW) | -2.6251 | 0.6915 | -0.6054 | -0.2012 |

Table 6. Comparison between three algorithms on averages of KL-divergences

Fig. 10 shows accuracy comparison between LW and HMP-GMR for the four CLG BNs. The X axis denotes KL-divergence for HMP-GMR. The Y axis denotes KL-divergence for LW. For CLG BN 1, HMP-GMR provided much better accuracy than LW. For CLG BN 2, LW provided better accuracy than HMP-GMR. For CLG BN 3, HMP-GMR provided better accuracy than LW. For CLG BN 4, LW and HMP-GMR performed similarly, but as can be seen in Fig.9, the results for LW were more variable.

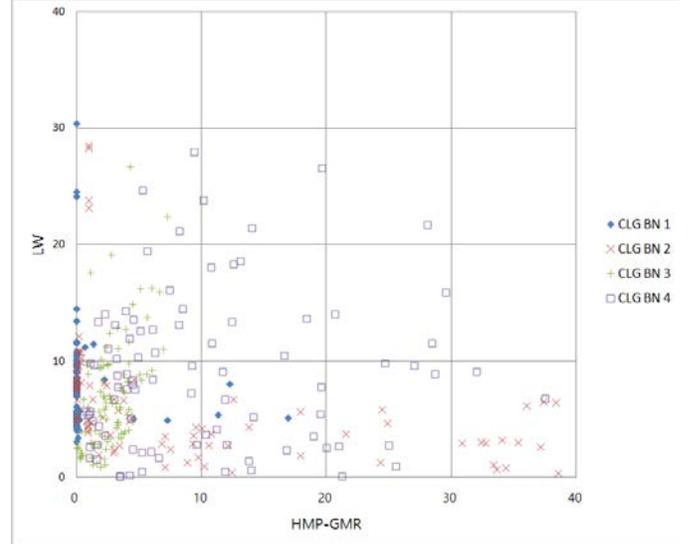

Fig. 10. Accuracy Comparison between LW and HMP-GMR for four CLG BNs

Accuracy from HMP-GMR was lower in comparison with LW, for the CLG BN containing many loops. CLG BN 2 contained 501 cycles, while CLG BNs 1 and 4 contained 21 cycles and CLG BN 3 contained no cycles. More extensive investigations would be needed to determine whether this superiority of LW over HMP-GMR generalizes to arbitrary BNs with many loops. Accuracies from HMP-GMR did not depend much on the number of discrete states. CLG BN 1 contained four discrete states, while the CLG BN 3 contained 16384 configurations of the discrete states. For these networks, HMP-GMR provided better accuracy than LW regardless of how many discrete states a CLG BN contains.

We can consider whether to use HMP-GMR or LW according to the features of the CLG BN. The following list shows suggestions in which HMP-GMR can be chosen or not in terms of the number of configurations of the discrete states and the number of cycles.

1. **Small number of configurations of the discrete states and small number of cycles**

    In this case, our experiment showed better accuracy from HMP-GMR in comparison with LW under a given time restriction. LW requires many samples to improve accuracy, while HMP-GMR uses message passing approach, which can provide exact results for a polytree network [Pearl, 1988]. In a simple-topology network (i.e., small number of configurations of the discrete states and small number of cycles), when

HMP-GMR converges in time, it can provide high accuracy.

2. **Small number of configurations of the discrete states and large number of cycles**

   When there are many cycles, HMP-GMR can diverge. This behavior depends on the network topology, the placement of continuous and discrete nodes, the conditional distributions, and the pattern of evidence. When divergence occurs, HMP-GMR halts during message passing because of numeric underflow. A feature to detect when the pi and lambda messages are going out of bounds and stop the algorithm may be useful, but intermediate results from before the algorithm diverges are of doubtful usefulness. In the case of a large number of cycles, LW can perform better than HMP-GMR.

3. **Large number of configurations of the discrete states and small number of cycles**

   HMP-GMR reduces the maximum allowable number of components, which is influenced by the number of configurations of the discrete states. So, HMP-GMR can tolerate many numbers of configurations of the discrete states, while LW requires many samples for many numbers of configurations of the discrete states. In this case, we observed that for the four CLG BNs, HMP-GMR could converge and provide good accuracy.

4. **Large number of configurations of the discrete states and large number of cycles**

   Although LW can tolerate many cycles, it can perform poorly when the necessary number of samples to achieve good accuracy is too large for the available time limit. Also, HMP-GMR can perform poorly because of many cycles. In this case, the better choice between LW and HMP-GMR may vary depending on specific features of the problem. For example, when the number of configurations of the discrete states is relatively smaller than the number of cycles, LW can be used. When the number of configurations of the discrete states is relatively larger than the number of cycles, HMP-GMR can be used. However, in any case, accuracies for both approaches may be low.

## VI. CONCLUSION

We have developed an extended message passing algorithm for CG hybrid Bayesian networks to overcome the exponential growth in components of the Gaussian mixture model. Our experiments demonstrated scalability, accuracy, and optimal settings for the complex hybrid BNs. In our experiments, both the original hybrid message passing inference and the hybrid junction tree inference showed exponential growth in execution time. The Gaussian mixture reduction method presented in this paper addressed this problem. Another issue we should address was to define ways to choose the optimal settings to achieve desired accuracy and execution time. For this, we presented a preprocessing algorithm to optimize the maximum allowable number of components and the optimal maximum iteration. The algorithm enables HMP-GMR to provide better accuracy within a predefined time. For the four CLG BNs we investigated, we observed that the accuracy from using a single Gaussian component was nearly as good as the setting found by our optimization method, and the difference in accuracy was not statistically significant. Note that other networks with complex topologies, very unlikely evidence configurations, and/or deterministic or near-deterministic relationships might be different. We observed that the accuracy results for a loopy CG BN were less than for a poly CG BN. To address this, we can use a clustering inference (e.g., Hybrid-JT) with Gaussian mixture reduction. These issues will be addressed in future work.

ACKNOWLEDGMENTS

The research was partially supported by the Office of Naval Research (ONR), under Contract#: N00173-09-C-4008. We appreciate Dr. W. Sun and Dr. K. C. Chang for the initial research regarding HMP-GMR.

REFERENCES

[1] Lerner, U., & Parr, R. (2001, August). Inference in hybrid networks: Theoretical limits and practical algorithms. In Proceedings of the Seventeenth conference on Uncertainty in artificial intelligence (pp. 310-318). Morgan Kaufmann Publishers Inc..

[2] Lauritzen, S. L. (1992). Propagation of probabilities, means, and variances in mixed graphical association models. Journal of the American Statistical Association, 87(420), 1098-1108.

[3] Lauritzen, S. L., & Jensen, F. (2001). Stable local computation with conditional Gaussian distributions. Statistics and Computing, 11(2), 191-203.

[4] Lerner, U. N. (2002). Hybrid Bayesian Networks for Reasoning about Complex Systems. Ph.D. dissertation, Stanford University, Stanford, CA.

[5] Sun, W., Chang, K. C, & Laskey, K. B. (2010). Scalable Inference for Hybrid Bayesian Networks with Full Density Estimation. Proceedings of the 13th International Conference on Information Fusion.

[6] Sun, W., & Chang, K. C. (2010, April). Direct message passing for hybrid Bayesian networks and performance analysis. In SPIE Defense, Security, and Sensing (pp. 76970S-76970S). International Society for Optics and Photonics.

[7] Sun, W. (2007). Efficient Inference for Hybrid Bayesian Networks. PhD Dissertation. George Mason University.

[8] Pearl, J. (1988). Probabilistic Reasoning in Intelligent Systems: Networks of Plausible Inference. San Mateo, CA, USA: Morgan Kaufmann Publishers.

[9] Salmond, D. J. (1990). Mixture reduction algorithms for target tracking in clutter. In OE/LASE'90, 14-19 Jan., Los Angeles, CA (pp. 434-445). International Society for Optics and Photonics.

[10] West, M. (1993). Approximating posterior distributions by mixture. Journal of the Royal Statistical Society. Series B (Methodological).

[11] Williams, J. L. (2003). Gaussian mixture reduction for tracking multiple maneuvering targets in clutter (no. afit/ge/eng/03-19). air force inst of tech wright-patterson afb oh school of engineering and management.

[12] Williams, J. L., & Maybeck, P. S. (2003). Cost-function-based Gaussian mixture reduction for target tracking. In Proceedings of the Sixth International Conference of Information Fusion (Vol. 2, pp. 1047-1054).

[13] Schrempf, O. C., Feiermann, O., & Hanebeck, U. D. (2005, July). Optimal mixture approximation of the product of mixtures. In Information Fusion, 2005 8th International Conference on (Vol. 1, pp. 8-pp). IEEE.

[14] Runnalls, A. R. (2007). Kullback-Leibler approach to Gaussian mixture reduction. Aerospace and Electronic Systems, IEEE Transactions on, 43(3), 989-999.

[15] Kullback, S., & Leibler, R. A. (1951). On information and sufficiency. The annals of mathematical statistics, 79-86.

[16] Chen, H. D., Chang, K. C., & Smith, C. (2010, April). Constraint optimized weight adaptation for gaussian mixture reduction. In SPIE Defense, Security, and Sensing (pp. 76970N-76970N). International Society for Optics and Photonics.

[17] Chang, K. C., & Sun, W. (2010, December). Scalable fusion with mixture distributions in sensor networks. In Control Automation Robotics & Vision (ICARCV), 2010 11th International Conference on (pp. 1251-1256). IEEE.

[18] Uhlmann, J. K. (1995). Dynamic map building and localization: New theoretical foundations (Doctoral dissertation, University of Oxford).

[19] Park, C. Y., Laskey, K. B., Costa, P. C. G., & Matsumoto, S. (2015). Message Passing for Hybrid Bayesian Networks using Gaussian Mixture Reduction. In Digital Information Management (ICDIM), 2015 Tenth International Conference on , vol., no., pp.210-216.

[20] Kozlov, A. V., & Koller, D. (1997, August). Nonuniform dynamic discretization in hybrid networks. In Proceedings of the Thirteenth conference on Uncertainty in artificial intelligence (pp. 314-325). Morgan Kaufmann Publishers Inc..

[21] Fung, R., & Chang, K. C. (1989). Weighting and integrating evidence for stochastic simulation in Bayesian networks. In UAI.

[22] Henrion, M. (1988). Propagation of uncertainty in Bayesian networks by probabilistic logic sampling. In Proc. UAI.

[23] Shenoy, P. P. (2006). Inference in hybrid Bayesian networks using mixtures of Gaussians. Uncertainty in Artificial Intelligence: Proceedings of the Twenty-Second Conference (UAI-06).

[24] Gilks, W. R., Richardson, S., & Spiegelhalter, D. J. (1996). Introducing markov chain monte carlo. Markov chain Monte Carlo in practice, 1, 19.

[25] Bidyuk, B., & Dechter, R. (2007). Cutset Sampling for Bayesian Networks. J. Artif. Intell. Res.(JAIR), 28, 1-48.

[26] Murphy, K. P., Weiss, Y., & Jordan, M. I. (1999, July). Loopy belief propagation for approximate inference: An empirical study. In Proceedings of the Fifteenth conference on Uncertainty in artificial intelligence (pp. 467-475). Morgan Kaufmann Publishers Inc..

[27] Johnson, D. B. (1975). Finding all the elementary circuits of a directed graph. SIAM Journal on Computing, 4(1), 77-84.

[28] Chen, C. H. (2010). Stochastic simulation optimization: an optimal computing budget allocation (Vol. 1). World scientific.